\documentclass[letterpaper]{article} 

\usepackage{adjustbox}
\usepackage[table]{xcolor}

\usepackage{aaai2026}  
\usepackage{times}  
\usepackage{helvet}  
\usepackage{courier}  
\usepackage[hyphens]{url}  
\usepackage{graphicx} 
\usepackage{natbib}  
\usepackage{caption} 

\usepackage{algorithm}

\usepackage{newfloat}
\usepackage{listings}
\DeclareCaptionStyle{ruled}{labelfont=normalfont,labelsep=colon,strut=off} 
\floatstyle{ruled}
\newfloat{listing}{tb}{lst}{}
\floatname{listing}{Listing}
\usepackage{amsfonts}
\usepackage{amsmath}
\usepackage{amssymb}
\usepackage[hyphens]{url}
\usepackage{url}
\usepackage{xcolor}
\usepackage{cleveref}
\usepackage[subrefformat=parens,labelformat=parens]{subfig}
\usepackage{threeparttable,booktabs}
\usepackage{makecell}
\usepackage{multirow}
\usepackage[]{algpseudocode}                               
\algrenewcommand\textproc{\texttt}
\makeatletter\let\float@addtolists\relax\makeatother
\usepackage{algorithm}
\usepackage{pgfplots}
\usepackage{pgfplotstable}
\pgfplotsset{compat=1.18}
\usepackage{courier}
\usepackage[inline]{enumitem}
\usepackage{tabularx}
\usepackage{soul}
\usepackage{bm}
\usepackage{pifont}                                        
\usepackage{bbding}                                        
\usepackage[super]{nth}

\newcommand{\minisection}[1]{\noindent{\textbf{#1}.}}

\newcommand{\simfunc}{\text{sim}}

\newtheorem{myproblem}{\textbf{Problem}}

\newtheorem{mydefinition}{\textbf{Definition}}
\newtheorem{mytheorem}{\textbf{Theorem}}
\newtheorem{myproof}{\textbf{Proof}}
\newtheorem{mycorollary}{\textbf{Corollary}}

\graphicspath{{./figs/}}

\title{KCLNet: Electrically Equivalence-Oriented Graph Representation Learning for Analog Circuits}

\author{%
    Peng Xu\equalcontrib\textsuperscript{\rm 1}, \
    Yapeng Li\equalcontrib\textsuperscript{\rm 2}, \
    Tinghuan Chen\textsuperscript{\rm 2}, \
    Tsung-Yi Ho\textsuperscript{\rm 1},  \
    Bei Yu\textsuperscript{\rm 1}
}

\affiliations {
    \textsuperscript{\rm 1} Department of Computer Science \& Engineering, The Chinese University of Hong Kong\\
    \textsuperscript{\rm 2} School of Science and Engineering, The Chinese University of Hong Kong (Shenzhen)
}

\begin{document}

\maketitle
\pagestyle{empty}

\begin{abstract}
\label{sec:abs}
Digital circsuit representation learning has made remarkable progress in the electronic design automation domain, effectively supporting critical tasks such as testability analysis and logic reasoning.
However, representation learning for analog circuits remains challenging due to their continuous electrical characteristics compared to the discrete states of digital circuits. 
This paper presents a direct current (DC) electrically equivalent-oriented analog representation learning framework, named \textbf{KCLNet}.
It cosmprises an asynchronous graph neural network structure with electrically-simulated message passing and a representation learning method inspired by Kirchhoff's Current Law (KCL).
This method maintains the orderliness of the circuit embedding space by enforcing the equality of the sum of outgoing and incoming current embeddings at each depth, which significantly enhances the generalization ability of circuit embeddings.
KCLNet offers a novel and effective solution for analog circuit representation learning with electrical constraints preserved.
Experimental results demonstrate that our method achieves significant performance in a variety of downstream tasks, e.g., analog circuit classification, subcircuit detection, and circuit edit distance prediction.
\end{abstract}

\section{Introduction}
\label{sec:intro}

\begin{figure}[!t]
    \centering
    \includegraphics[width=\linewidth]{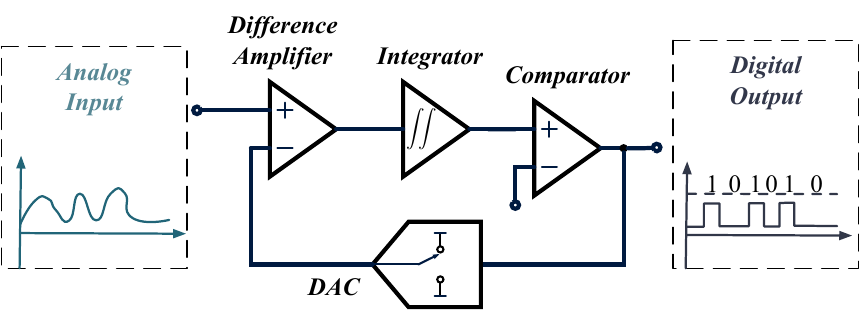}
    \caption{$\Delta\Sigma$ analog-to-digital converter~(ADC), a typical analog circuit type.
    }
    \label{fig:analog_circuits}
\end{figure}

Analog and mixed-signal electronic systems have become the backbone of modern technological advancement, driving innovations from medical instrumentation to autonomous vehicles. 
Their pervasive presence relies fundamentally on the seamless integration of analog and digital circuits.
Within this electronic design paradigm, analog circuits emerge as the critical interface bridging the physical and digital worlds \citep{gray2009analysis}. 
As shown in \Cref{fig:analog_circuits}, analog circuit components such as operational amplifiers and data converters translate real-world signals—temperature variations, sound waves, or wireless transmissions—into digitally processable information while maintaining signal fidelity. 
Despite occupying less than 20\% of modern system-on-chip (SoC) area, analog blocks determine over 40\% of product reliability and power efficiency metrics \citep{sansen2007analog}. 
The design of analog circuits confronts challenges rooted in their sensitivity to physical characteristics.
Unlike digital counterparts operating in well-defined Boolean domains, analog components are perpetually exposed to parasitic, process variation, and layout-dependent effects.
Thus, analog circuit design, from schematic to layout, remains primarily a manual, time-consuming, and error-prone task \citep{xu2024performancesurvey} at present.
The inherent electrical complexities of analog circuits create heavy reliance on human expertise, forming a critical bottleneck in automation.

Emerging machine learning paradigms, particularly graph neural networks (GNNs) \citep{gnn2020_Hamilton, 2021GNNs}, exhibit substantial potential in decoding the design complexity inherent to analog circuits. 
Naturally, the analog circuits can be viewed as graphs that consist of devices and nets, and their connections.
By converting analog circuits into graphs, contemporary methodologies have demonstrated significant progress across critical analog design sub-tasks \citep{chen2021universal, gao2021layout, kunal2020gana, settaluri2020autockt, wang2020gcn, dongcktgnn, hou2024cktgen, tu2025smart, xu2024paroute, xu2025paroute2}.
For analog topology classification, graph-level embeddings are utilized to retrieve the targeted layout templates of analog circuits, such as amplifiers and filters in \citep{kunal2020gana}.
For subgraph identification, parallel advancements in layout synthesis leverage subgraph-aware GNN architectures to reduce symmetry violations compared to manual design as in \citep{chen2021universal, gao2021layout}.
At the transistor level, sizing optimization frameworks employing node-centric GNNs reduce simulation iterations \citep{settaluri2020autockt, wang2020gcn, dongcktgnn, hou2024cktgen}.
Additionally, analog circuit embeddings are also employed in various stages of analog circuit design, including large-scale subgraph matching during testing and performance optimization during routing \citep{tu2025smart, xu2024paroute, xu2025paroute2}.

Despite widespread use of task-specific circuit embeddings, general analog circuit representation learning remains deficient in physical prior integration and lacks dedicated research as a standalone field.
Some efforts attempt to develop multi-modality analog pretraining frameworks, where name-based text embeddings and auxiliary layout data are incorporated to incorporate domain knowledge \citep{ren2020paragraph, zhu2022tag}.
While these multi-modality frameworks have demonstrated capability in the analog layout aspect,  they have not demonstrated general representation capability at the analog circuit level.
As the counterpart of analog circuits, digital circuit representation has demonstrated significant potential in leveraging physical laws, with Boolean algebra operations \cite{jonsson1988relation} becoming a cornerstone of digital circuit representation learning \cite{li2022deepgate, shi2023deepgate2, shi2024deepgate3, zheng2025deepgate, wang2024fgnn2, wu2025circuit}.
The intrinsic physical characteristics of analog devices are still neglected.
Incorporating physical laws into analog circuit representation learning is crucial to match the success achieved in digital representation learning.



To address these limitations, we propose an electrical physics-inspired representation learning framework, named \textbf{KCLNet}, which incorporates fundamental electrical principles through \textit{Kirchhoff's current law (KCL)}.
Kirchhoff's Current Law states that the algebraic sum of all currents flowing into a node (junction) is equal to the algebraic sum of all currents flowing out \citep{paul2001fundamentals, rewienski2011perspective, athavale2018kirchoff}.
It is a fundamental principle in electrical circuit theory that forms the cornerstone of classical circuit analysis.
The KCL is integrated through an electrically simulated message passing and a novel contrastive learning scheme.
Our contributions are summarized as follows:
\begin{itemize}
    \item To honor the electrical current flow, we convert analog circuits to directed acyclic graphs (DAGs) and propose an asynchronous message passing scheme with layer-wise propagation from voltage to ground nodes.
    \item Based on that, a physics-informed contrastive objective is designed where depth-wise embeddings enforce Kirchhoff’s current conservation positives and node masking creates electrically inconsistent negatives.
    We theoretically justify that the proposed contrastive objective can preserve the electrical principle of KCL.
    \item The experimental results show that the analog circuit embeddings learned by our proposed KCLNet can benefit a variety of downstream tasks, with 20.77\% improvement in Acc@1 gain in analog circuit classification, 43.36\% mAP gain in analog subcircuit detection, and 1.6$\%$ MAE gain in analog circuit graph-edit-distance prediction.
\end{itemize}

\begin{links} 
Our codes are available at \link{Code}{https://github.com/shipxu123/KCLNet}.
\end{links}

\begin{table}[!t]
    \centering
    \begin{tabular}{|c|c|c|}
        \hline
        Device & Number of Pins & Pin types \\ \hline\hline
        NMOS     &   $n_d + n_g + n_s + n_b$ & nd, ng, ns, nb \\
        PMOS     & $n_d + n_g + n_s + n_b$ & pd, pg, ps, pb \\
        NPN      & $n_b + n_c + n_e$       & nb, nc, ne     \\
        PNP      & $n_b + n_c + n_e$       & nb, nc, ne     \\
        Diode    & 2                       & n+, n-         \\
        Resistor & 2                       & n+, n-         \\
        Capacitor& 2                       & n+, n-         \\
        Inductor & 2                       & n+, n-         \\ \hline
    \end{tabular}
    \caption{Typical Analog Device Types}
    \label{tab:deive_pin_types}
\end{table}

\section{Related Work and Preliminaries}
\subsection{Graph Neural Networks (GNNs)}

GNNs consist of two major components, where the $\mathrm{aggregation}$ step aggregates node features of target nodes' neighbors, and the $\mathrm{combination}$ step passes the previous aggregated features to networks to generate node embeddings.
Mathematically, we can update node $v$'s embedding at the $l$-th layer by: 
\begin{equation}
\label{eq:GNN}
\begin{aligned}
    \boldsymbol{e}_{v}^{l} &= \mathrm{AGG}\left( \left\{\boldsymbol{h}_{u}^{l-1} | \forall u \in \mathcal{N}(v)\right\}\right),\\
    \boldsymbol{h}_{v}^{l} &= \mathrm{COMBINE}\left(\boldsymbol{h}_{v}^{l-1},  \boldsymbol{e}_{v}^{l}\right),
\end{aligned}
\end{equation}
where $\mathcal{N}(v)$ denotes the neighbours of $v$.

\subsection{Analog Representation Learning (ARL)}
Analog circuits pose unique representation challenges due to their bipartite structure (devices and nets) and heterogeneous device types.
Recent work explores GNNs to learn analog circuit representations directly from graph structure \cite{kunal2020gana, settaluri2020autockt, wang2020gcn, dongcktgnn, hou2024cktgen}.
We give the formal definition of an analog circuit graph as follows:

\begin{mydefinition}[Device]
Each device $v_d \in \mathcal{V}_d$ is associated with attributes such as type (NMOS, RES, CAP, etc.), parameters (e.g., W, L, resistance, capacitance, etc.), and connectivity information, as shown in \Cref{tab:deive_pin_types}.
\end{mydefinition}

\begin{mydefinition}[Net] 
Each net $v_n \in \mathcal{V}_n$ is a junction where multiple devices connect, with topological metrics decided by the connected pins of devices as shown in \Cref{tab:deive_pin_types}.
\end{mydefinition}



\begin{mydefinition}[Analog Circuit Graph]
The analog circuit graph consists of two groups of nodes $\mathcal{V}_{d}, \mathcal{V}_{n}$, corresponding to devices and nets.
Those two groups of nodes are connected by a set of edges $\mathcal{E}$, where different edge types correspond to different pin types, presenting connection information.
Hence, the analog circuit graph has a form of bipartite graph as $\mathcal{G} = \{\mathcal{V}_{d}, \mathcal{V}_{n}, \mathcal{E}\}$.
\end{mydefinition}

\begin{figure*}[!tb]
    \centering
    \includegraphics[width=.85\linewidth]{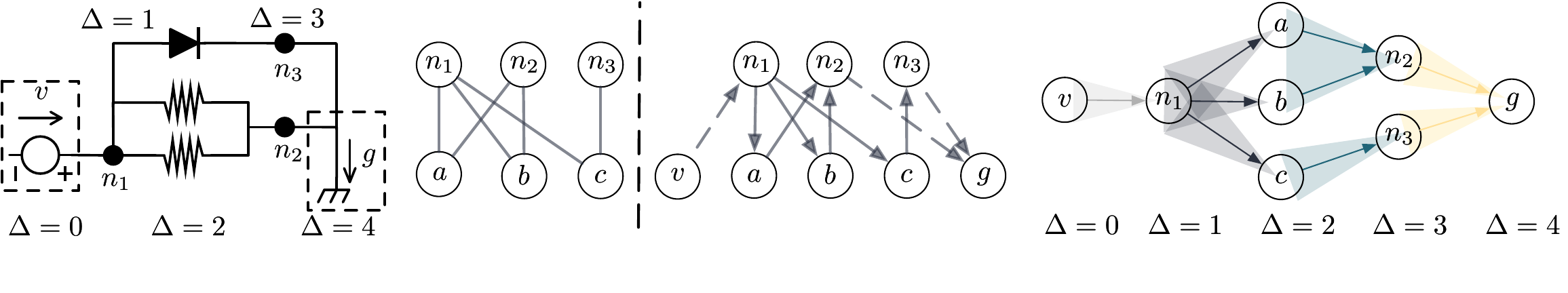}
    \caption{Directed acyclic circuit graph representation: (1) analog circuit; (2) convert bipartite graph representation (left) to DAG via topology sorting  (right); (3) electrically-simulated asynchronous message passing scheme.
    }
    \label{fig:agnn}
\end{figure*}

\section{Method}
\subsection{Electrically-Simulated Async Message Passing}

\minisection{Directed Acyclic Circuit Graph Conversion}
In traditional analog representation learning, existing methods frequently disregard the directions of the analog circuit graph \citep{chen2021universal, kunal2020gana, tu2025smart}.
This omission leads to neglecting current flow directionality during the learning process, resulting in a loss of crucial physical information.
Although behavioral-level analog circuits are used to construct directed graphs in \citep{dongcktgnn} and \citep{hou2024cktgen}, this approach is specifically tailored for specific analog devices, e.g., operational transconductance amplifier (OTA), rather than being general-purpose.
Notably, in analog circuits, current typically originates from the power supply and traverses along the signal path to the ground, as noticed in \citep{gao2021layout}.
Consequently, to incorporate the physical characteristics of current flow into representation learning, this paper introduces a novel approach by designating the \textit{voltage} and \textit{ground} nodes as special nodes. 
The original analog circuit graph, a bipartite graph, is thus transformed into a directed acyclic graph (DAG) next.

We first discuss the conversion from undirected to directed graphs via topology sorting~\cite{kahn1962topological}.
The \textit{voltage} nodes and the \textit{ground} nodes are added as the start nodes and the end nodes.
Based on the current flow direction in the circuit, assign directions to each edge in $\mathcal{E}$.
Typically, current flows from voltage nodes to the ground nodes, so the edge directions should align with this current flow.
Given that the analog circuit graph is a bipartite graph with edges only between devices and nets, we make the following theorem by topologically traversing from voltage nodes to ground nodes:

\begin{mytheorem}[Alicyclic Guarantee after Conversion]
The original graph is a bipartite graph where edges only exist between devices and nets.
Assume voltage and ground nodes are special devices added to the graph:
1. Voltage nodes have only outgoing edges with connected nets;
2. Ground nodes have only incoming edges with connected nets.
The converted directed graph will be acyclic by traversing the bipartite graph via topology sorting, with the voltage and ground nodes as the start and end points.
\end{mytheorem}

The proof of this theorem is provided in Appendix A.6.
Following the theorem, the resulting graph becomes a DAG after conversion with voltage nodes as the start nodes and ground nodes as the terminal nodes, as shown in \Cref{fig:agnn}.
All paths in this DAG, pointing from voltage nodes to ground nodes, are consistent with the physical characteristics and signal flow of the circuit.
This conversion process formalizes the structure of an analog circuit into a DAG, providing a foundation graph format for subsequent analog circuit representation learning.

    
\begin{figure*}[!tb]
    \centering
    \includegraphics[width=.85\linewidth]{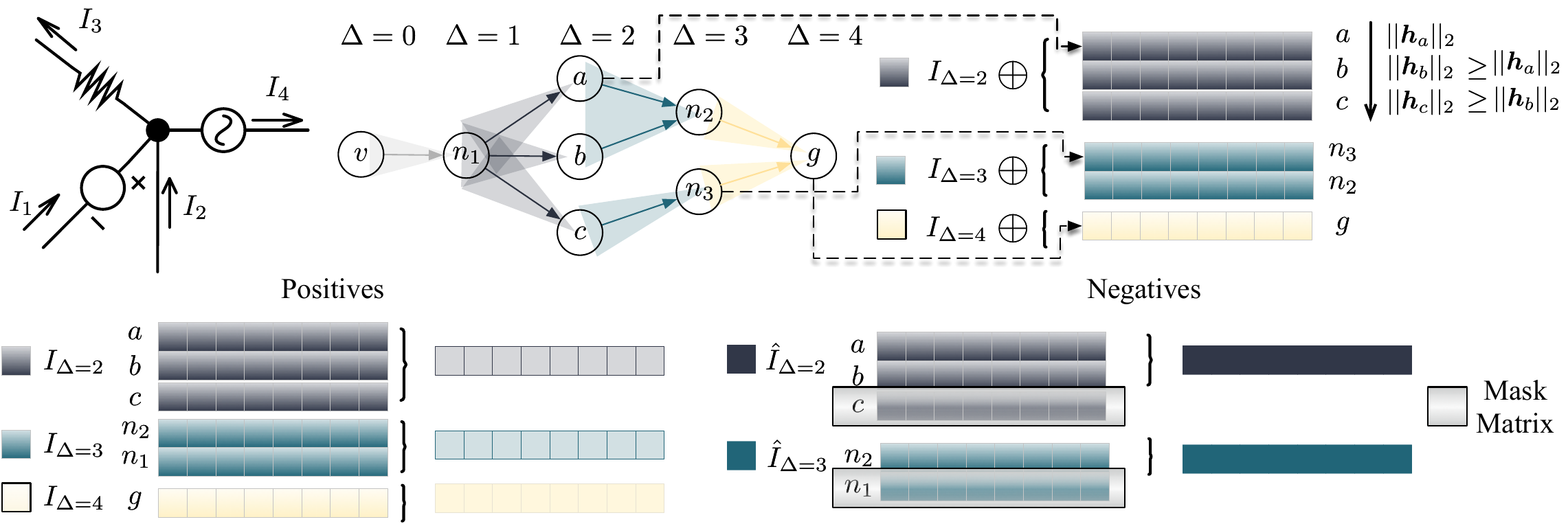}
    \caption{
        The framework of the physics-guided contrastive learning scheme, named the KCL Loss
    }
    \label{fig:kcl_framework}
\end{figure*}

\minisection{Depthwise Message Passing Scheme}
Conventional graph neural networks (GNNs) operate synchronously as illustrated in \citep{bruna2014spectral,defferrard2017convolutional,hamilton2018inductive,Petar2018graph,xu2019powerful}.
In \emph{synchronous message passing}, all messages flow simultaneously along edges during each iteration. 

For better capturing current flow direction as in \citep{dimo1975nodal, wedepohl2002modified}, we propose an \textbf{asynchronous GNN~(AGNN)} architecture that simulates depthwise message passing from voltage nodes to ground nodes.
We take \textit{voltage nodes} as the root nodes of the first layer through topological ordering while fixing \textit{ground nodes} as the terminal layer nodes.
The electrically-simulated message passing process initiates from the voltage source node at depth~$0$, propagates sequentially from depth~$1$ to~$d$, and ultimately reaches the \textit{ground nodes}.
At each depth level, only the vertices that have received messages from the previous depth propagate messages to their direct successors. 
The \Cref{fig:agnn} demonstrates an example of embedding nodes using this asynchronous GNN. 
Formally, for a target vertex $v$, the aggregation scheme of depth-asynchronous GNN at the $\Vec{\Delta}$-th depth can be described as:
\begin{equation}
 \begin{aligned}
& \Vec{e}_{\{\Vec{i}: \mathcal{D}(\Vec{i}, \Vec{v})=\Vec{\Delta}-\Vec{l}\}}^{(l)}=\operatorname{AGG}\left(\left\{\Vec{h}_{u}^{(k-1)}: u \in \mathcal{N}(i)\right\}\right), \\
& \Vec{h}_{\{\Vec{i}: \mathcal{D}(\Vec{i}, \Vec{v})=\Vec{\Delta}-\Vec{l}\}}^{(l)}=\operatorname{COMBINE}\left(\Vec{e}_i^{(l)}, \Vec{h}_i^{(\Vec{0})}\right),
\end{aligned}   
\end{equation}
where $\mathcal{D}(i, v)$ denotes the distance between vertices $i$ and $v$ in the graph, and $\bm{h}_i^{(0)}$ represents the initial feature vector of vertex $i$.
We used the same $\operatorname{AGG}$ and $\operatorname{COMBINE}$ functions as GCN, with the sum of normalized neighbor embeddings in our implementation~\cite{kipf2017semisupervised}. 
Thus, our AGNN can be regarded as a GCN variant.
Specifically, during the $k$-th iteration of a depth-$\Delta$, aggregation occurs exclusively at vertices located at a distance of $\Delta - k$ from the root node $v$, which selectively integrates messages from their topological predecessors.
This aggregated output is then fused with the vertex's original features to generate its updated representation vector.

\subsection{Electrically Equivalent Contrastive Learning}
\label{subsec:kcl_loss}

\minisection{Kirchhoff's Current Law Analysis}
KCL is one of the most fundamental theorems in analog circuit analysis \citep{paul2001fundamentals, rewienski2011perspective, athavale2018kirchoff}. It describes a crucial principle: the total current entering a node always equals the total current exiting it, as shown in 
\Cref{fig:kcl_framework}.
Formally, it can be defined as:
\begin{equation}
    I_{in1} + I_{in2} +\cdots \Leftrightarrow I_{out1}+I_{out2}+\cdots.
\end{equation}

A key extension of the KCL is that the algebraic sum of input and output currents of multiple circuit components, i.e., a supernode, remains equal~\cite{Zeng2021, Zhu2024}, which is frequently applied in the analysis of the current values at different depths.
Our key idea is to preserve this equivalence in the analog circuit embedding space at different depths:

\begin{equation}
    \label{eq:kcl_embedding}
    \begin{aligned}
          \Vec{I}_{\Vec{\Delta}} &= \Vec{I}_{\Vec{\Delta^{\prime}}},\\
          \sum \Vec{h}_{\{\Vec{i}: \mathcal{D}(\Vec{i}, \Vec{v})=\Vec{\Delta}-\Vec{l}\}}^{(l)} &= \sum \Vec{h}_{\{\Vec{j}: \mathcal{D}(\Vec{j}, \Vec{v})=\Vec{\Delta^{\prime}} -\Vec{l}\}}^{(l)},\\
    \end{aligned}
\end{equation}
where $\Vec{\Delta}$ and $\Vec{\Delta^{\prime}}$ are different depths of current analog circuit graph. 
The electrical equivalence relation "$\Leftrightarrow$" is an equivalence relation under the constraint of \Cref{eq:kcl_embedding}.
For all device sets within one equivalent depth, the sum of the embeddings of all devices they consist of should be equal.
This equation finds a natural equivalence that exists in universal analog circuits with the same current inputs and outputs.


\begin{figure*}[!tb]
    \centering
    \centering\includegraphics[width=.85\linewidth]{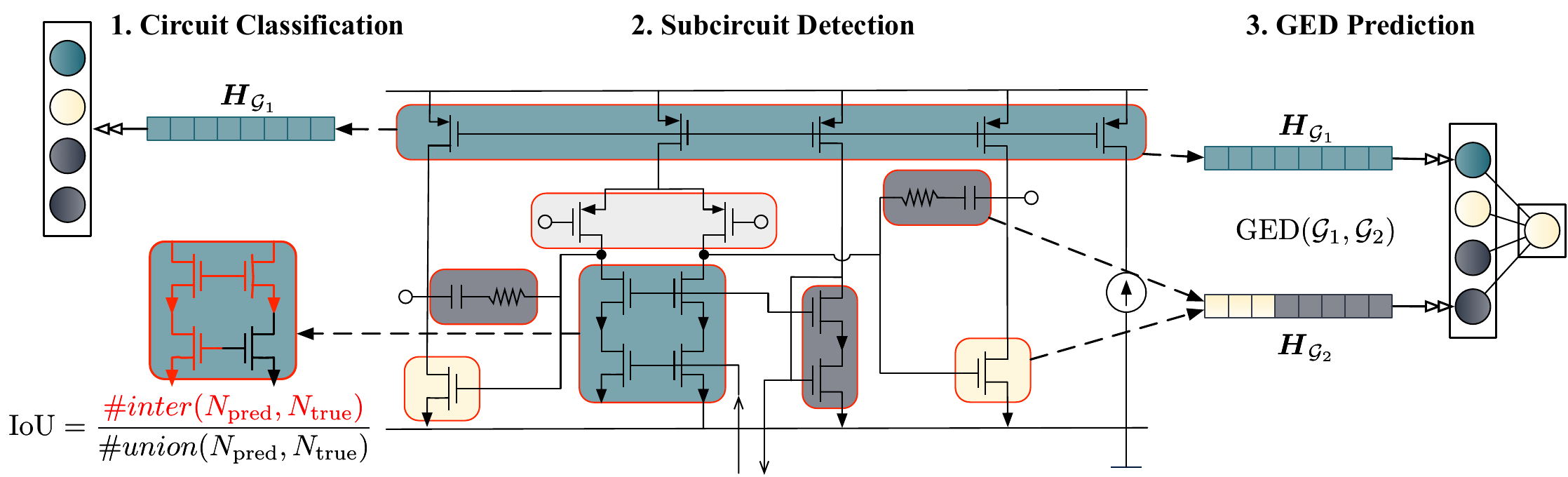}
    \caption{The illustration of the downstream tasks: (1) \texttt{Analog circuit classification}; (2) \texttt{Analog subcircuit detection}; (3) \texttt{Analog GED preidction}.}
    \label{fig:subtasks}
\end{figure*}

\minisection{Electrical Equivalence as Positives}
For an analog circuit DAG graph $\mathcal{G}_{i}=\left\{\mathcal{V}_d^{i}, \mathcal{V}_n^{i}, \mathcal{E}^{i}\right\}$, we first use the GNN encoder to process all nodes from different depths in this graph and get their embeddings. 
As shown in \Cref{fig:kcl_framework}, the subcircuit embedding pairs $\left(\Vec{I}_{\Vec{\Delta}},  \Vec{I}_{\Vec{\Delta^{\prime}}}\right)$ from different depths are treated as positive pairs, whose embedding discrepancy will be minimized. 
According to \Cref{eq:kcl_embedding}, a straightforward loss function is therefore:
$$
\begin{aligned}
L & =  \sum_{{\Vec{\Delta^{\prime}}} \neq\Vec{\Delta}}^{d} \simfunc( \Vec{I}_{\Vec{\Delta}} , \Vec{I}_{\Vec{\Delta^{\prime}}} ),\\
\end{aligned}
$$
where $\operatorname{sim}\left(\Vec{v}_1, \Vec{v}_2\right)$ is the cosine similarity function $\frac{\Vec{v}_1^{\top} \Vec{v}_2}{\left\|\Vec{v}_1\right\| \cdot\left\|\Vec{v}_2\right\|}$ as in \citep{he2020momentum}, and $d$ is the maximum depth.

This simple constraint is critical for enhancing the quality of analog circuit embeddings. We show that by enforcing such constraints, the trained neural network can thus preserve the electrical characteristic of KCL with the following theorem:
\begin{mytheorem}[Kirchhoff's Current Law preservation]
Let $\{\Vec{I}_{\Delta_1}, \Vec{I}_{\Delta_2}, \ldots, \Vec{I}_{\Delta_n}\}$ and $\{\Vec{I}_{\Delta_1'}, \Vec{I}_{\Delta_2'}, \ldots, \Vec{I}_{\Delta_n'}\}$ be two sets of vectors in $\mathbb{R}^d$, where $n \leq d$ and $0 \leq \Delta_i, \Delta_i' < d$, and $d$ is the longest distance of the the analog circuit graph.
Then, there exists a non-trivial linear map $\phi: \mathbb{R}^d \to \mathbb{R}$ such that $\phi(\Vec{I}_{\Delta_i}) = \phi(\Vec{I}_{\Delta_i'})$ for all $i = 1, 2, \ldots, n$. 
\end{mytheorem}

The proof of this theorem is provided in Appendix A.7.
The theorem shows that for the current embeddings of different layers, there always exists a nonlinear mapping function that determines the input and output current values, with the algebraic sum satisfying KCL.
Based on that, we also have the following corollary
, with the proof also attached in Appendix A.7:

\begin{mycorollary}
Suppose that for each $i$, the distance between the corresponding vectors after normalization is sufficiently small, i.e., 
$ 1 - \hat{\Vec{I}}_{\Delta_i}^{\top}\hat{\Vec{I}}_{\Delta_i^{\prime}} \leq \epsilon$.
A smaller $\epsilon$ makes constructing the desired map $\phi(\cdot)$ easier.
\end{mycorollary}

Directly minimizing the loss function is ineffective, as the model would become degenerate by producing all-zero vectors for every subcircuit embedding \citep{chen2020simple, pang2022unsupervised}.
Typical approaches to address this involve using negative sampling or contrastive learning techniques.
In our work, we introduce a novel technique for generating negative samples by leveraging the electrical contradiction of KCL.

\minisection{Electrical Contradiction as Hard Negatives}
We introduce a novel technique for generating negative samples $ \hat{\Vec{I}}_{\Vec{\Delta}}$ by selectively applying dropout to node embeddings at each depth $\Vec{\Delta}$.
Because the sum of incoming currents at each layer is equal, artificially creating an imbalance by discarding node embeddings with higher current values naturally generates negative samples from the circuit's perspective. 

As shown in \Cref{fig:kcl_framework}, compute the squared $L_2$ norm of its embedding $\Vec{h}_{\{\Vec{i}: \mathcal{D}(\Vec{i}, \Vec{v})}$ as a measure of its magnitude and define a binary mask to identify top-$k$ largest nodes: 
\begin{equation}
\begin{aligned}
    \Vec{r}_i^{\Vec{\Delta}} &= \left\| \Vec{h}_{\{\Vec{i}: \mathcal{D}(\Vec{i}, \Vec{v})=\Vec{\Delta}-\Vec{l}\}}^{(l)} \right\|_2,\\
    \Vec{M}^{\Vec{\Delta}}[i] &= 
    \begin{cases} 
        1 & \text{if } r_i^{\Vec{\Delta}} \in \text{top-}k \text{ of } \{r_1^{\Vec{\Delta}}, \dots, r_n^{\Vec{\Delta}}\}\\
        0 & \text{otherwise}
    \end{cases}.
\end{aligned}
\end{equation}

Apply the dropout by element-wise multiplication of the original embeddings with the inverted mask:
\begin{equation}
    \hat{\Vec{I}}_{\Vec{\Delta}}= \sum \left(\Vec{h}_{\{\Vec{i}: \mathcal{D}(\Vec{i}, \Vec{v})\}} \odot \left( \Vec{1} - \Vec{M}^{\Vec{\Delta}} \right)\right),
\end{equation}
where $ \odot $ denotes element-wise multiplication, and $\Vec{1}$ is a matrix of ones. This zeros out embeddings for the top-$k$ nodes, creating hard negatives.

\minisection{KCL Loss}
Combine the positives and negatives inspired by KCL, we follow the contrastive framework in \citep{you2020graph, Zhu2020vf} to derive the KCL Loss, the training objective for $\left(\Vec{I}_{\Vec{\Delta}}, \Vec{I}_{\Vec{\Delta^{\prime}}}\right)$ with $N = d \times (d - 1)$ pairs is:
$$
\mathcal{L} = -\log \frac{e^{\operatorname{sim}\left(\Vec{I}_{\Vec{\Delta}}, \Vec{I}_{\Vec{\Delta^{\prime}}}\right) / \tau}}{\sum_{j=1}^N e^{\operatorname{sim}\left(\Vec{I}_{\Vec{\Delta}}, \hat{\Vec{I}}_{\Vec{\Delta}^{\prime}}\right) / \tau}},
$$
where $\tau$ is a temperature hyperparameter.

\begin{table*}[!t]
    \centering
    \begin{tabular}{@{}cccccc@{}}
    \toprule
    Method & Acc@1$\uparrow$ & Acc@2$\uparrow$ & Acc@5$\uparrow$ & Recall$\uparrow$ & F1 Score$\uparrow$ \\ 
    \midrule
    \multicolumn{6}{@{}l}{\textbf{Base Models}} \\
    \midrule
    GCN       & 0.561$_{\pm{\text{0.061}}}$ & 0.784$_{\pm{\text{0.040}}}$ & 0.913$_{\pm{\text{0.014}}}$ & 0.405$_{\pm{\text{0.087}}}$ & 0.404$_{\pm{\text{0.121}}}$  \\
    GAT       & 0.479$_{\pm{\text{0.105}}}$ & 0.681$_{\pm{\text{0.148}}}$ & 0.794$_{\pm{\text{0.118}}}$ & 0.327$_{\pm{\text{0.060}}}$ & 0.275$_{\pm{\text{0.069}}}$  \\
    GATv2     & 0.498$_{\pm{\text{0.100}}}$ & 0.694$_{\pm{\text{0.139}}}$ & 0.799$_{\pm{\text{0.133}}}$ & 0.322$_{\pm{\text{0.051}}}$ & 0.270$_{\pm{\text{0.075}}}$  \\ 
    
    GIN       & 0.786$_{\pm{\text{0.077}}}$ & 0.883$_{\pm{\text{0.058}}}$ & 0.931$_{\pm{\text{0.039}}}$ & 0.648$_{\pm{\text{0.101}}}$ & 0.669$_{\pm{\text{0.118}}}$  \\
    SAGE & 0.774$_{\pm{\text{0.019}}}$ & 0.927$_{\pm{\text{0.023}}}$ & 0.938$_{\pm{\text{0.004}}}$ & 0.626$_{\pm{\text{0.028}}}$ & 0.638$_{\pm{\text{0.055}}}$  \\
    \midrule
    \multicolumn{6}{@{}l}{\textbf{GraphCL Variants}} \\
    \midrule
    GraphCL$_{\text{GCN}}$       & 0.679$_{\pm{\text{0.131}}}$ & 0.795$_{\pm{\text{0.120}}}$ & 0.910$_{\pm{\text{0.063}}}$ & 0.558$_{\pm{\text{0.150}}}$ & 0.564$_{\pm{\text{0.186}}}$  \\
    GraphCL$_{\text{GAT}}$       & 0.537$_{\pm{\text{0.108}}}$ & 0.794$_{\pm{\text{0.046}}}$ & 0.919$_{\pm{\text{0.046}}}$ & 0.408$_{\pm{\text{0.135}}}$ & 0.368$_{\pm{\text{0.173}}}$  \\
    GraphCL$_{\text{GATv2}}$     & 0.493$_{\pm{\text{0.112}}}$ & 0.650$_{\pm{\text{0.075}}}$ & 0.809$_{\pm{\text{0.112}}}$ & 0.401$_{\pm{\text{0.104}}}$ & 0.357$_{\pm{\text{0.148}}}$  \\ 
    GraphCL$_{\text{GIN}}$       & 0.875$_{\pm{\text{0.083}}}$ & 0.943$_{\pm{\text{0.009}}}$ & 0.950$_{\pm{\text{0.006}}}$ & 0.741$_{\pm{\text{0.083}}}$ & 0.759$_{\pm{\text{0.102}}}$ \\
    
    GraphCL$_{\text{SAGE}}$      & 0.899$_{\pm{\text{0.013}}}$ & 0.958$_{\pm{\text{0.001}}}$ & 0.962$_{\pm{\text{0.001}}}$ & 0.726$_{\pm{\text{0.024}}}$ & 0.744$_{\pm{\text{0.036}}}$  \\
    \midrule
    \multicolumn{6}{@{}l}{\textbf{Our Methods}} \\
    \midrule
    
    \textbf{KCLNet} & \textbf{0.949}$_{\pm\text{0.004}}$ & \textbf{0.958}$_{\pm\text{0.005}}$ & \textbf{0.964}$_{\pm\text{0.004}}$ & \textbf{0.829}$_{\pm\text{0.015}}$ & \textbf{0.853}$_{\pm\text{0.011}}$ \\
    $w.o$ Pos$^{\text{KCL}}$ & 0.938$_{\pm{\text{0.005}}}$ & 0.942$_{\pm{\text{0.006}}}$ & 0.944$_{\pm{\text{0.003}}}$ & 0.794$_{\pm{\text{0.013}}}$ & 0.830$_{\pm{\text{0.012}}}$  \\
    $w.o$ Neg$^{\text{KCL}}$ & 0.939$_{\pm{\text{0.004}}}$ & 0.946$_{\pm{\text{0.010}}}$ & 0.949$_{\pm{\text{0.010}}}$ & 0.801$_{\pm{\text{0.015}}}$ & 0.682$_{\pm{\text{0.335}}}$  \\
    $w.o$ Pos+Neg$^{\text{KCL}}$ & 0.938$_{\pm{\text{0.003}}}$ & 0.941$_{\pm{\text{0.005}}}$ & 0.944$_{\pm{\text{0.002}}}$ & 0.794$_{\pm{\text{0.009}}}$ & 0.822$_{\pm{\text{0.018}}}$  \\
    \bottomrule
    \end{tabular}
    \caption{Performance comparison on analog circuit classification. Top performers in each category are {bold}.}
    \label{tab:main_results_cls}
\end{table*}

\begin{table*}[!t]
    \centering
    \begin{tabular}{@{}cccccc@{}}
    \toprule
    Method & mAP$\uparrow$ & Recall$\uparrow$ & F1 Score$\uparrow$ & AUC$\uparrow$ & IoU$\uparrow$ \\ 
    \midrule
    \multicolumn{6}{@{}l}{\textbf{Base Models}} \\
    \midrule
    GCN       & 0.368$_{\pm{\text{0.002}}}$ & 0.260$_{\pm{\text{0.004}}}$ & 0.225$_{\pm{\text{0.003}}}$ & 0.796$_{\pm{\text{0.002}}}$ & 0.144$_{\pm{\text{0.002}}}$ \\
    GAT       & 0.383$_{\pm{\text{0.013}}}$ & 0.290$_{\pm{\text{0.033}}}$ & 0.262$_{\pm{\text{0.038}}}$ & 0.808$_{\pm{\text{0.012}}}$ & 0.170$_{\pm{\text{0.032}}}$ \\
    GATv2     & 0.382$_{\pm{\text{0.017}}}$ & 0.303$_{\pm{\text{0.059}}}$ & 0.279$_{\pm{\text{0.072}}}$ & 0.807$_{\pm{\text{0.022}}}$ & 0.183$_{\pm{\text{0.051}}}$ \\
    
    GIN       & 0.553$_{\pm{\text{0.029}}}$ & 0.667$_{\pm{\text{0.019}}}$ & 0.695$_{\pm{\text{0.020}}}$ & 0.932$_{\pm{\text{0.005}}}$ & 0.565$_{\pm{\text{0.021}}}$ \\
    SAGE      & 0.355$_{\pm{\text{0.004}}}$ & 0.255$_{\pm{\text{0.002}}}$ & 0.211$_{\pm{\text{0.002}}}$ & 0.785$_{\pm{\text{0.001}}}$ & 0.133$_{\pm{\text{0.001}}}$ \\
    %
    \midrule
    \multicolumn{6}{@{}l}{\textbf{GraphCL Variants}} \\
    \midrule
    GraphCL$_{\text{GCN}}$  & 0.363$_{\pm{\text{0.014}}}$ & 0.222$_{\pm{\text{0.005}}}$ & 0.181$_{\pm{\text{0.008}}}$ & 0.762$_{\pm{\text{0.008}}}$ & 0.115$_{\pm{\text{0.005}}}$ \\
    GraphCL$_{\text{GAT}}$  & 0.371$_{\pm{\text{0.016}}}$ & 0.231$_{\pm{\text{0.027}}}$ & 0.193$_{\pm{\text{0.032}}}$ & 0.757$_{\pm{\text{0.032}}}$ & 0.121$_{\pm{\text{0.023}}}$ \\
    GraphCL$_{\text{GATv2}}$& 0.389$_{\pm{\text{0.010}}}$ & 0.271$_{\pm{\text{0.096}}}$ & 0.236$_{\pm{\text{0.105}}}$ & 0.779$_{\pm{\text{0.052}}}$ & 0.154$_{\pm{\text{0.077}}}$ \\
    
    GraphCL$_{\text{GIN}}$  & 0.434$_{\pm{\text{0.033}}}$ & 0.539$_{\pm{\text{0.076}}}$ & 0.539$_{\pm{\text{0.090}}}$ & 0.898$_{\pm{\text{0.017}}}$ & 0.420$_{\pm{\text{0.070}}}$ \\
    GraphCL$_{\text{SAGE}}$ & 0.361$_{\pm{\text{0.016}}}$ & 0.253$_{\pm{\text{0.015}}}$ & 0.213$_{\pm{\text{0.013}}}$ & 0.776$_{\pm{\text{0.014}}}$ & 0.134$_{\pm{\text{0.009}}}$ \\
    \midrule
    \multicolumn{6}{@{}l}{\textbf{Our Methods}} \\
    \midrule
    
    \textbf{KCLNet} & \textbf{0.622$_{\pm{\text{0.002}}}$} & \textbf{0.721$_{\pm{\text{0.002}}}$} & \textbf{0.753$_{\pm{\text{0.002}}}$} & \textbf{0.949$_{\pm{\text{0.000}}}$} & \textbf{0.634$_{\pm{\text{0.003}}}$} \\
    $w.o$ Pos$^{\text{KCL}}$ & 0.374$_{\pm{\text{0.027}}}$ & 0.399$_{\pm{\text{0.098}}}$ & 0.381$_{\pm{\text{0.117}}}$ & 0.845$_{\pm{\text{0.042}}}$ & 0.273$_{\pm{\text{0.096}}}$ \\
    $w.o$ Neg$^{\text{KCL}}$ & 0.602$_{\pm{\text{0.039}}}$ & 0.706$_{\pm{\text{0.040}}}$ & 0.736$_{\pm{\text{0.046}}}$ & 0.944$_{\pm{\text{0.010}}}$ & 0.615$_{\pm{\text{0.053}}}$ \\
    $w.o$ Pos+Neg$^{\text{KCL}}$ & {0.561}$_{\pm{\text{0.035}}}$ & {0.654}$_{\pm{\text{0.038}}}$ & {0.674}$_{\pm{\text{0.045}}}$ & {0.931}$_{\pm{\text{0.010}}}$ & {0.545}$_{\pm{\text{0.051}}}$ \\
    \bottomrule
    \end{tabular}
    \caption{Performance comparison on subcircuit detection task.}
    \label{tab:main_results_det}
\end{table*}

\section{Experiments}
\label{sec:exp}
We conduct experiments to address the following issues:
\textbf{(1)} \textit{How does KCLNet compare to general GNNs and graph pre-trained methods in terms of performance?}
\textbf{(2)} \textit{How effective is the proposed KCL Loss?}
\textbf{(3)} \textit{What is the impact of each KCLNet module on its overall performance?}

\subsection{Experimental Setup}
\minisection{Dataset and Downstream Tasks}
We employ the analog circuits dataset \texttt{ANALOG} in \citep{tu2025smart} generated using the analog circuit topology synthesis framework \citep{zhao2022analog}.
The generated circuits contain fundamental topologies ranging from basic building blocks to complex industrial-scale systems. 
The details of the used analog circuit topologies are in Appendix A.2.

In this paper, we introduce three downstream applications as shown in \Cref{fig:subtasks} to verify our pre-trained circuit embedding.
Three datasets for sub-tasks were developed for analog circuit-related tasks.
For analog circuit classification, the \texttt{ANALOG} dataset was expanded by adjusting sizing parameters, resulting in \texttt{ANALOG-CLS-428k}; for analog subcircuit detection, following Kunal et al.’s ~\citep{kunal2020gana} framework,  base circuit categories were classified via manual annotation, creating \texttt{ANALOG-DET-242k} with 242,320 samples; for analog graph edit distance (GED) prediction, \texttt{ANALOG-GED-194k} was generated via mutations on existing data. 
All datasets emphasize balanced splits and proportional representation of circuit characteristics to ensure robust model evaluation.
The details of the definition of subtasks and the dataset partition scheme are provided in Appendix A.2.

\minisection{Evaluation Metrics}
For the analog circuit classification task, we adopt standard classification metrics for evaluation: top-k accuracy (Acc@1, Acc@2, Acc@5), True Positive Rate (Recall), and F1 Score.
For the analog subcircuit detection task, we use common evaluation metrics in detection tasks, including mAP (mean Average Precision), Recall, F1 Score, and IoU (Intersection over Union).
We utilize common evaluation metrics in regression tasks for the graph edit distance prediction task, including MAE and MSE.
The reported results are the average of the best results in $5$ runs from different random seeds.

\minisection{Baseline Methods}
The baseline methods we compare include two categories:
(1) mainstream GNN encoders without pretraining;
(2) mainstream GNN representation learning method combined with different GNN Encoders.
For the mainstream GNN Encoders, we adopted the following commonly used GNN frameworks as our analog circuits encoder: 
GCN~\cite{kipf2017semisupervised},
GAT~\cite{Petar2018graph}, GIN~\cite{xu2019powerful}, Graphsage~\cite{hamilton2018inductive}, 
GAT\_v2~\cite{brodyattentive}.
For mainstream GNN pretraining methods, 
we utilized the general graph pre-training method named GraphCL~\cite{you2020graph}
, combined with different graph encoders as comparison methods.
All methods are compared based on the same hyperparameter.
The baseline methods' implementation details are provided in Appendix A.4.

\subsection{Experimental Results}
\label{subsec:exp_results}

\minisection{Analog Circuits Classification} 
As shown in \Cref{tab:main_results_cls}, we comprehensively compared the experimental results of various methods for the analog circuit classification task.
Specifically, KCLNet achieves the highest accuracy among all baseline models, showcasing its superior representation ability for downstream classification.
Compared to the pre-trained graph model GraphCL$_{\text{GCN}}$, our method brings about a 39.76\% and 51.19\% improvement in Acc@1 and F1 Score, respectively.
When compared to the best performing methods, GraphCL$_{\text{GIN}}$ and GIN, it offers a 5.56\% and 20.77\% improvement in Acc@1 and F1 Score, respectively.
These results confirm that KCLNet has better representation compared with existing approaches in the analog circuit classification task.

\minisection{Analog Subcircuits Detection}
As shown in \Cref{tab:main_results_det}, KCLNet also outperforms other methods in the subcircuit detection task.
Compared to the best performing methods, GraphCL$_{\text{GIN}}$ and GIN, KCLNet achieves improvements of up to 43.36\%, 33.80\%, 39.61\%, 5.67\%, and 51.02\% in the mAP, Recall, F1 Score, AUC, and IoU metrics, respectively.
It can be observed that the general-purpose graph pre-training algorithm GraphCL$_{\text{GIN}}$ fails to enhance the model's performance on downstream circuit-related tasks compared to GIN.
The key issue is that these generic pre-training tasks often fail to capture equivalence in circuit diagrams.
These improvements in metrics further indicate that KCLNet's representation learning capabilities enhance the detection accuracy, recall, and robustness.

\begin{figure}[tb!]
    \centering
    \includegraphics[width=.65\linewidth]{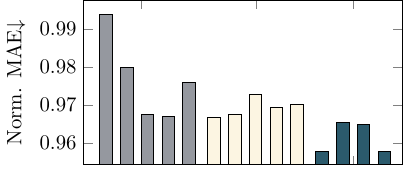}\\
    \includegraphics[width=.65\linewidth]{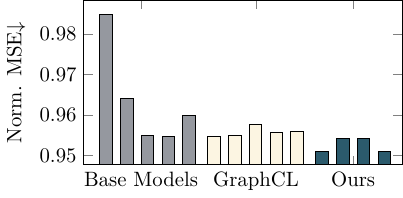}
    \caption{The averaged and normalized performance comparison on the analog circuit GED prediction task.}
    \label{fig:ged_exp}
\end{figure}

\minisection{Graph-Edit-Distance Prediction}
As shown in \Cref{fig:ged_exp}, the proposed contrastive learning framework reduces up to 3.8\% MAE compared to vanilla GNNs. Additionally, compared with methods that have experienced improvement through general pretraining, such as GraphCL$_{\text{GIN}}$, KCLNet maintains a similar improvement with up to 1.6\%. It can also be observed that general graph pretraining methods do not necessarily guarantee an improvement in the performance of GCN, which can be seen from the close MAE and MSE values.

\begin{figure}[tb!]
    \centering
    \subfloat[]{\includegraphics[height=2.65cm]{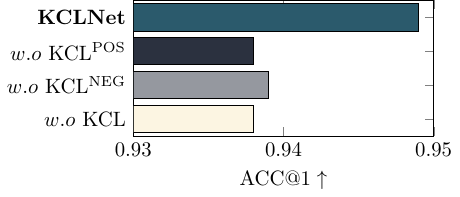}\label{fig:ablation_kcl_a}}\\
    \subfloat[]{\includegraphics[height=2.65cm]{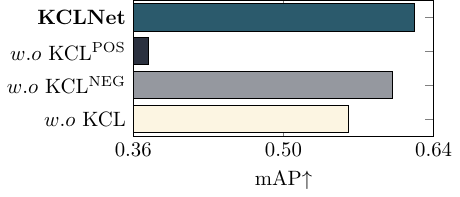}\label{fig:ablation_kcl_b}}
    \caption{Comparison between KCLNet with different variants of KCL Loss.}
    \label{fig:ablation_kcl_combined}
\end{figure}

\subsection{Ablation Studies}
\label{subsec:ablation_exp}

\minisection{KCL-Inspired Loss at Work}
A notable advantage of KCLNet lies in its ability to generate positive embeddings in the analog circuits from an electrical perspective.
To verify this, \Cref{fig:ablation_kcl_a} and \Cref{fig:ablation_kcl_b} show performance comparisons of different KCL Loss variants.
The {\textit{w.o \text{KCL}$^{\text{POS}}$}} uses general graph-augmented positive samples to maximize alignment, {\textit{w.o \text{KCL}$^{\text{NEG}}$}} replaces KCL-based negatives with different graph samples in the same batch, and {\textit{w.o $\text{KCL}$}} removes KCL Loss entirely, relying solely on asynchronous message passing.
The results show that KCL Loss is crucial for final performance, with positive embeddings being the most important.

\minisection{Asynchronous Message Passing Scheme}
In \Cref{table:agnn}, we compare the performance differences between the proposed AGNN and synchronous GNN.
AGNN outperforms the best synchronous model, GIN, by 19.6\% and 22.9\% in Acc@1 and F1-score in the cls task.
It also improves the mAP by 1.4\% in the det task, surpassing all other GNN models. 
These results confirm the effectiveness of electrically-simulated asynchronous message passing.

\begin{figure}[tb!]
    \centering
    \includegraphics[height=2.75cm]{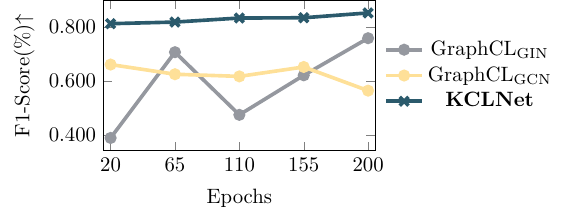}
    \caption{The pre-training epochs impact of KCLNet on classification, which outperforms general graph pretraining at earlier epochs.}
    \label{fig:ablation_epochs}
\end{figure}

\minisection{The Effect of Training Epochs in KCLNet}
As shown in \Cref{fig:ablation_epochs}, KCLNet surpasses GraphCL pretrained methods for different epochs (GraphCL$_{\text{GIN}}$ and GraphCL$_{\text{GCN}}$) starting from $20$ epochs, showing strong early performance. 
Unlike other frameworks that rely on augmentation for positive samples with more pretraining epochs needed, KCLNet uses KCL's physical prior knowledge to generate positive embeddings without augmentation, reducing the needed pretraining epochs.

\begin{table}[tb!]
    \centering
    \resizebox{.98\linewidth}{!}
    {
        \begin{tabular}{c|cc|cc}
            \toprule
            \multirow{1}{*}{} &
            \multicolumn{2}{c|}{Classification} &
            \multicolumn{2}{c}{Detection}
            \\
            ~ & {Acc@1$\uparrow$} & {F1-Score$\uparrow$}  & {mAP$\uparrow$}  & {IoU$\uparrow$}\\
            \midrule
            \multirow{1}{*}{GCN}   & 0.561$_{\pm{\text{0.061}}}$ & 0.404$_{\pm{\text{0.121}}}$ & 0.368$_{\pm{\text{0.002}}}$ & 0.144$_{\pm{\text{0.002}}}$ \\
            \multirow{1}{*}{GAT}   & 0.496$_{\pm{\text{0.127}}}$ & 0.306$_{\pm{\text{0.101}}}$ & 0.383$_{\pm{\text{0.013}}}$ & 0.170$_{\pm{\text{0.032}}}$ \\
            \multirow{1}{*}{GATv2} & 0.498$_{\pm{\text{0.100}}}$ & 0.270$_{\pm{\text{0.075}}}$ & 0.382$_{\pm{\text{0.017}}}$ & 0.183$_{\pm{\text{0.051}}}$ \\
            \multirow{1}{*}{GIN}   & 0.786$_{\pm{\text{0.077}}}$ & 0.669$_{\pm{\text{0.118}}}$ & 0.553$_{\pm{\text{0.029}}}$ & \textbf{0.565$_{\pm{\text{0.021}}}$} \\
            \multirow{1}{*}{SAGE}  & 0.774$_{\pm{\text{0.019}}}$ & 0.638$_{\pm{\text{0.055}}}$ & 0.355$_{\pm{\text{0.004}}}$ & 0.133$_{\pm{\text{0.001}}}$ \\
            \midrule
            \multirow{1}{*}{AGNN} & \textbf{0.938$_{\pm{\text{0.003}}}$} & \textbf{0.822$_{\pm{\text{0.018}}}$} &  \textbf{{0.561}$_{\pm{\text{0.035}}}$} & {0.545}$_{\pm{\text{0.051}}}$ \\
            \bottomrule
        \end{tabular}
    }
    \captionof{table}{Performance comparison between synchronous and asynchronous message passing~(AGNN).}
    \label{table:agnn}
\end{table}

\section{Conclusion}
\label{sec:conclu}

In this work, we use an electrically-simulated asynchronous graph neural network as the analog circuit encoder, leveraging Kirchhoff's current law to aid representation learning by enforcing current embedding conservation across depths. Experiments show the model learns vital physical priors, significantly enhancing generalization across analog sub-tasks. 
We propose three future directions: 1) modeling Kirchhoff's voltage law from a voltage perspective; 2) exploring better encoders like graph transformers; 3) incorporating additional inputs like SPICE codes to assist learning.


\section{Acknowledgments}
This work is supported by The National Key Research and Development Program of China (No.~2023YFB4402900), The National Natural Science Foundation of China (No.~92573108 and No.~62304197),
and The Research Grants Council of Hong Kong SAR (No.~CUHK14211824 and No.~CUHK14201624).

\clearpage
{
\IfFileExists{main.bbl}{

}{
\bibliography{ref/Top-sim, ref/all}
}
}


\clearpage
\clearpage
\appendix
\setcounter{secnumdepth}{2}

\section{Appendix}
\subsection{The Introduction of GNNs}
\label{subsec:appendix_rw_gnns}
Graph neural networks (GNNs) are powerful representation learning techniques \cite{xu2019powerful} with many key applications \cite{hamilton2018inductive}. 
Early GNNs are motivated from the spectral perspective, such as Spectral GNN \cite{bruna2014spectral} that applies the Laplacian operators directly. 
ChebNet~\citep{defferrard2017convolutional} approximates these operators using summation instead to avoid a high computational cost. 
GCN~\citep{kipf2017semisupervised} further simplifies ChebNet by using its first order, and reaches the balance between efficiency and effectiveness, revealing the message-passing mechanism of modern GNNs.
Concretely, recent GNNs aggregate node features from neighbors and stack multiple layers to capture long-range dependencies.
For instance, GraphSAGE~\citep{hamilton2018inductive} concatenates node features with mean/max/LSTM pooled neighboring information.
GAT~\citep{Petar2018graph} aggregates neighbor information using learnable attention weights.
GIN~\citep{xu2019powerful} converts the aggregation as a learnable function based on the Weisfeiler-Lehman test instead of prefixed ones, aiming to maximize the power of GNNs.

GNNs consist of two major components, where the $\operatorname{aggregation}$ step aggregates node features of target nodes' neighbors, and the $\operatorname{combination}$ step passes the previous aggregated features to networks to generate node embeddings.
Mathematically, we can update node $v$'s embedding at the $l$-th layer by: 
\begin{equation}
\label{eq:GNN_appendix}
\begin{aligned}
    \Vec{e}_{v}^{l} &= \operatorname{AGGREGATE}\left( \left\{\Vec{h}_{u}^{l-1} | \forall u \in \mathcal{N}(v)\right\}\right),\\
    \Vec{h}_{v}^{l} &= \operatorname{COMBINE}\left(\Vec{h}_{v}^{l-1},  \Vec{e}_{v}^{l}\right),
\end{aligned}
\end{equation}
where $\mathcal{N}(v)$ denotes the neighbours of $v$.

\begin{table*}[!t]
\caption{Integrated circuit dataset statistics across different circuit categories.}
\label{tab:circuit_dataset}
    \centering
    \begin{adjustbox}{width=.80\textwidth}
    \begin{tabular}{@{}llccccccc@{}}
    \toprule
    \rowcolor{gray!10}
    \textbf{Category} & \textbf{Circuit Examples} & \textbf{\#NMOS} & \textbf{\#PMOS} & \textbf{\#RES} & \textbf{\#CAP} & \textbf{\#Nets} & \textbf{\#Nodes} & \textbf{\#Edges} \\ 
    \midrule
    \multicolumn{9}{@{}l}{\textbf{Data Converters} \small{Signal conversion between analog and digital domains, precision-critical}} \\
    \midrule
     & DAC\_R2R\_Type & 34 & 34 & 103 & 0 & 131 & 307 & 466 \\
     & ADC\_SAR\_Small & 27 & 27 & 0 & 0 & 36 & 93 & 196 \\
     & ADC\_SAR\_Large & 27 & 27 & 0 & 0 & 36 & 93 & 196 \\
    \midrule
    \multicolumn{9}{@{}l}{\textbf{Digital Logic} \small{Discrete-valued binary operations, high noise immunity, well-defined states}} \\
    \midrule
     & Decoder\_3to7 & 38 & 38 & 0 & 0 & 44 & 123 & 274 \\
     & Digital\_Mux & 35 & 35 & 0 & 0 & 43 & 115 & 223 \\
     & ResNetwork\_PNP & 0 & 0 & 80 & 0 & 80 & 163 & 161 \\
     & SignBit\_Detector & 27 & 27 & 4 & 0 & 32 & 93 & 208 \\
    \midrule
    \multicolumn{9}{@{}l}{\textbf{Signal Circuits} \small{Continuous signal processing, high linearity, sensitive to noise}} \\
    \midrule
    Power Management & LDO\_Regular & 18 & 18 & 33 & 0 & 54 & 126 & 199 \\
    Current Reference & CurrentRef\_5uA & 24 & 29 & 0 & 0 & 35 & 91 & 183 \\
    Op Amp & OpAmp\_FirstStage & 17 & 25 & 0 & 0 & 28 & 73 & 142 \\
    Resistor Divider & ResistorDivider & 63 & 63 & 60 & 0 & 72 & 261 & 578 \\
    Bandgap Reference & BandgapRef\_Mux & 20 & 20 & 0 & 0 & 26 & 69 & 147 \\
    \midrule
    \multicolumn{9}{@{}l}{\textbf{Timing and I/O} \small{Clock generation, signal synchronization, external interface management}} \\
    \midrule
    \multirow{2}{*}{Clock} & PLL\_PhaseDetector & 34 & 34 & 4 & 0 & 44 & 119 & 251 \\
     & PLL\_PostProcessor & 76 & 76 & 0 & 0 & 76 & 230 & 499 \\
    I/O Circuits & IO\_Buffer\_Pad & 19 & 25 & 5 & 0 & 31 & 83 & 159 \\
    Buffers & Buffer\_Calibrated & 50 & 76 & 1 & 1 & 77 & 208 & 441 \\
    \midrule
    \multicolumn{9}{@{}l}{\textbf{Other Components} \small{Design reuse, flexibility elements, standardized cell architecture}} \\
    \midrule
    Spare Cells & SpareCell\_Standard & 90 & 90 & 2 & 0 & 90 & 276 & 582 \\
    \bottomrule
    \end{tabular}
    \end{adjustbox}
\end{table*}

\subsection{Dataset Details}
\label{subsec:appendix_dataset_details}
\Cref{tab:circuit_dataset} provides the details for circuit topologies: transistor counts (PMOS/NMOS), passive components (\#C capacitors, \#R resistors), and node counts after graph conversion.
Our hierarchical benchmark structure enables comprehensive evaluation across different complexity tiers: 1) {small-scale:} matches core building blocks (inverters, push-pull stages) with complex implementations;
2) {medium-scale:} tests intermediate structures (operational amplifiers, output stages);
3) {large-scale:} evaluates system-level designs including ADC cores, and Industrial SerDes circuit ($>10,000$ nodes).
This framework spans circuit scales from $10^1$ to $10^3$ nodes, covering diverse analog circuit topologies.


\subsection{Downstream Tasks for Analog Representation Learning}
\label{subsec:appendix_downstream_tasks}
In this paper, we introduce three downstream applications to verify our pre-trained circuit embedding.

\begin{myproblem}[Analog Circuit Classification]
Assign a class label $y \in \{1, 2, \dots, C\}$ to a circuit graph $\mathcal{G}$.
Let $\mathcal{D}_{\text{cls}} = \{(G_i, y_i)\}_{i=1}^N$ be a dataset of labeled circuit graphs. The task is to learn a classifier $f_{\text{cls}}: \mathbb{R}^d \rightarrow \mathbb{R}^C$ such that:
$$
\hat{y} = \text{softmax}(f_{\text{cls}}(z)), \quad \text{where } z = \phi(G).
$$
The loss function is typically the cross-entropy function:
$$
\mathcal{L}_{\text{cls}} = -\sum_{i=1}^N \sum_{c=1}^C \mathbb{I}(y_i = c) \log \hat{y}_{i,c}.
$$
\end{myproblem}

\begin{myproblem}[Analog Subcircuit Detection.]
Let $\mathcal{D}_{\text{det}} = \{(G_i, S_i, v_{j}^i, y_{i,j})\}_{i=1}^M$, where $y_{i,j} \in \{0, 1\}$ indicates whether node $v_{j}^i$ in circuit $G_i$ belongs to subcircuit type $S_i$. The task is to learn a detector $f_{\text{det}}: \mathbb{R}^d \times \mathbb{R}^d \rightarrow [0, 1]$ such that:
$$
\hat{y}_{i,j} = \sigma(f_{\text{det}}(z_{v_j}, S_i)), \quad \text{where } z_{v_j} = \phi(v_j),
$$
where $\sigma$ is the sigmoid function.
The loss is cross-entropy:
$$
\mathcal{L}_{\text{det}} = -\sum_{i=1}^M \sum_{j=1}^{N_i} \left[ y_{i,j} \log \hat{y}_{i,j} + (1 - y_{i,j}) \log (1 - \hat{y}_{i,j}) \right],
$$
where $N_i$ is the number of nodes in circuit $G_i$.
\end{myproblem}

\begin{myproblem}[Analog Graph Edit Distance (GED) Prediction]
Predict the minimum number of edit operations (node/edge insertions, deletions, substitutions) Required to transform graph $G_1$ into $G_2$.
Let $\mathcal{D}_{\text{ged}} = \{(G_1^i, G_2^i, d_i)\}_{i=1}^K$, where $d_i$ is the ground-truth GED between $G_1^i$ and $G_2^i$. The task is to learn a regressor $f_{\text{ged}}: \mathbb{R}^d \times \mathbb{R}^d \rightarrow \mathbb{R}$ such that:
$$
\hat{d} = f_{\text{ged}}(z_1, z_2), \quad \text{where } z_1 = \phi(G_1), \, z_2 = \phi(G_2).
$$
The loss is mean squared error (MSE):
$$
\mathcal{L}_{\text{ged}} = \frac{1}{K} \sum_{i=1}^K (\hat{d}_i - d_i)^2.
$$
\end{myproblem}

\subsection{Dataset Partition Details}
\label{subsec:appendix_dataset_partition_details}

For the \texttt{analog circuit classification} task, we generated additional foundational subcircuit detection graphs and expanded the dataset scale by adjusting sizing parameters through further processing of the \texttt{ANALOG} dataset classified into $12$ general classes.
The dataset is partitioned into training,  validation, and testing sets of $360,000$, $28,800$, and $39,600$ analog circuit graphs, referred to as \texttt{ANALOG-CLS-428k}.
This split ensures a balanced representation of circuit types and complexities across all phases.

For the \texttt{analog subcircuit detection} task, we followed the experimental framework of Kunal~et~al. ~\citep{kunal2020gana}, classifying fundamental circuit elements into 12 distinct base categories through manual annotation. 
We adhere to the electrical characteristics of these components to integrate different circuit connections into a larger circuit.
The \textit{voltage} and \textit{ground} nodes are explicitly omitted from detection targets. The device nodes and net nodes retain their original classification labels.
The resulting dataset \texttt{ANALOG-DET-242k} contains $242,320$ samples distributed as training 76.9\%,  validation: 7.7\%, testing: 15.4\%.
This stratified partitioning Ensures the proportional representation of subcircuit categories and complexity levels across all splits, mitigating potential evaluation bias.

For the \texttt{analog graph edit distance (GED) prediction} task, we generated labeled training (70\%), validation (10\%), and test (20\%) datasets by performing mutating operations on the existing data.
We aim to predict the GED between two circuit graphs based on their embeddings. 
The purpose of this task is to demonstrate whether the learned embeddings of analog circuits can preserve the structural similarity between the circuits.
This ensures a balanced graph edit distance distribution across all phases. 
We named this dataset \texttt{ANALOG-GED-194k}.

\subsection{Experimental Implementation Details}
\label{subsec:appendix_impl}
All the experiments were performed on an Amazon EC2 cluster with an Intel Xeon E5-2686 v4 CPU, 61GB DRAM, and an NVIDIA Tesla V100 GPU

\minisection{The pretraining configuration.} 
In our experiments, we adopt $3$-layer GNN as the backbone.
Unless specified, our experiments follow the same settings for pretraining as GraphCL~\citep{you2020graph} :

\begin{itemize}
    \item $Graph Augmentation.$ We align the default augmentation strength $0.2$ with ~\citep{you2020graph}. We select the hyperparameter $\gamma$ controlling the trade-off in the optimization as $0.1$.
    \item $Optimizer.$ We use SGD to update models' parameters, i.e., $w$. The learning rate and SGD momentum are given as $0.025$ and $0$, respectively, where the learning rate has a cosine decay schedule for $200$ epochs. We fix the weight decay value, i.e., set $\rho_1=0.00125$.
    \item $Batch \ size.$ We adopt in-memory datasets, and the $batch\_size$ is fixed as $32$ for all pertaining methods. 
    \item $KCL\_Loss.$ In implementing KCL Loss of KCLNet, we set the $k$ in top-k to $1$ for negative samples, to facilitate handling layers with only one node. A layer-specific dropout strategy could be further explored in future work.
 \end{itemize}

At the downstream tasks, we adopt SGD as the optimizer and set the scheduler with cosine decay to adjust the learning rate. For fairness, the total number of epochs is fixed at $20$ for all GNN models. 
The details of the training configurations are listed as follows:
\begin{itemize}
\item \texttt{Analog circuit classification}:  We set the initial learning rate as $0.01$ and weight decay as $0.005$. The configuration for GNN models is as follows: $hidden\_size=64$, $dropout=0.6$, and using $ReLU$ as the activation function.
\item \texttt{Analog subcircuit detection}: We set the initial learning rate as $0.03$ and weight decay as $0.005$. The configuration for GNN models is as follows: $hidden\_size=64$, $dropout=0.6$, and using $ReLU$ as the activation function.
\item \texttt{Graph edit distance prediction}: We set the initial learning rate as $0.01$ and weight decay as $0.0025$. The configuration for GNN models is as follows: $hidden\_size=64$, $dropout=0.6$, and using $ReLU$ as the activation function.
\end{itemize}

\subsection{Proof for the Alicyclic Guarantee after Conversion theorem}
\label{subsec:proof_dag}
\begin{mytheorem}[Alicyclic Guarantee after Conversion]



Let $G = (V, E)$ be a directed graph constructed as follows:\\
\noindent (1) Node set $V = D \cup N \cup \{v\} \cup \{g\}$, where:
    \begin{itemize}
        \item $D$: Regular devices (resistors, capacitors, etc.);
        \item $N$: Nets;
        \item $\{v\}$: Voltage nodes (special devices);
        \item $\{g\}$: Ground nodes (special devices).
    \end{itemize}
(2) Edge set $E$ satisfies:
    \begin{itemize}
        \item $\forall d \in D \cup \{v\}$, there exists an edge $d \to n$ for each net $n \in N$ connected to $d$ (based on the original bipartite graph);
        \item No edges into $v$ or out of $g$.
    \end{itemize}

Let the \emph{original graph} be the undirected bipartite graph with node set $D \cup N$ and edges between devices and nets (excluding $\{v\}$ and $\{g\}$). If this original graph is bipartite between devices ($D$) and nets ($N$), then $G$ is acyclic.
\end{mytheorem}

\begin{myproof}
We prove acyclicity by defining a ranking function $\rho: V \to \mathbb{N}$ such that for every directed edge $(u, w) \in E$, $\rho(u) < \rho(w)$. This implies no cycles exist, as a cycle $u_1 \to u_2 \to \cdots \to u_k \to u_1$ would yield $\rho(u_1) < \rho(u_2) < \cdots < \rho(u_1)$, a contradiction.

\noindent
\textbf{Defining the ranking function $\rho$:}
Consider the undirected graph $H$ with node set $D \cup N \cup \{v\} \cup \{g\}$ and edges:
\begin{itemize}
    \item All edges from the original bipartite graph (between $D$ and $N$);
    \item Edges between $\{v\}$ and nets $n \in N$ connected to $v$;
    \item Edges between $\{g\}$ and all nets $n \in N$ connected to $g$.
\end{itemize}
$H$ is bipartite between devices ($D \cup \{v\} \cup \{g\}$) and nets ($N$) because:
\begin{itemize}
    \item The original graph is bipartite between $D$ and $N$ by assumption.
    \item Adding $\{v\}$ and $\{g\}$ to the device set preserves bipartiteness, as edges are only between devices and nets.
\end{itemize}
Perform a Breadth First Search (BFS) in $H$ starting from $v$ to assign layers:
\begin{align*}
    L_0 &= \{v\} \\
    L_1 &= \{ n \in N \mid n \text{adj to.} v \text{ in } H \} \\
    L_2 &= \{ d \in D \cup \{g\} \mid d \in \mathcal{N}(n), n \in L_1\} \\
    L_3 &= \{ n \in N \mid n \in \mathcal{N}(d), d \in L_2 \} \\
    &\vdots \\
    L_{\ell} &= \text{remaining nodes (until all are visited)}
\end{align*}
Since $H$ is bipartite, nodes alternate between nets (odd layers) and devices (even layers), and the layering is well-defined (no contradictions). Let $M = \max\{ \text{layer index} \}$ be the maximum layer number. Define $\rho(u)$ as:
\[
\rho(u) = 
\begin{cases} 
\text{layer}(u) & \text{if } u \notin \{g\} \\
M + 1 & \text{if } u \in \{g\}
\end{cases}
\]
Here, $\text{layer}(u)$ is the BFS layer index of $u$ in $H$. For $\{g\}$, we set $\rho(g) = M + 1$ to ensure it has the highest rank.

\noindent
\textbf{Verifying $\rho(u) < \rho(w)$ for all edges $(u, w) \in E$:}
Consider each type of edge in $G$:
\begin{itemize}
    \item \textbf{Edge } $d \to n$ \textbf{ for } $d \in D \cup \{v\}$, $n \in N$: \\
    By construction, $d$ and $n$ are adjacent in $H$, so $|\text{layer}(d) - \text{layer}(n)| = 1$. Since devices are in even layers and nets in odd layers:
    \begin{itemize}
        \item If $\text{layer}(d) < \text{layer}(n)$, then $\rho(d) = \text{layer}(d) < \text{layer}(n) = \rho(n)$.
        \item If $\text{layer}(d) > \text{layer}(n)$, then $\text{layer}(d) \geq \text{layer}(n) + 2$ (layers alternate), but $|\text{layer}(d) - \text{layer}(n)| = 1$, so this case is impossible. Thus, only $\text{layer}(d) < \text{layer}(n)$ occurs, implying $\rho(d) < \rho(n)$.
    \end{itemize}
    \item \textbf{Edge } $n \to g$ \textbf{ for } $n \in N$: \\
    Since $\rho(n) = \text{layer}(n) \leq M$ and $\rho(g) = M + 1$, we have $\rho(n) \leq M < M + 1 = \rho(g)$, so $\rho(n) < \rho(g)$.
\end{itemize}
Thus, for all edges $(u, w) \in E$, $\rho(u) < \rho(w)$.

\noindent
\textbf{Contradiction for cycles:}
Suppose, for contradiction, that $G$ contains a cycle $C = u_1 \to u_2 \to \cdots \to u_k \to u_1$. For each edge $u_i \to u_{i+1}$ (with $u_{k+1} = u_1$), $\rho(u_i) < \rho(u_{i+1})$. This implies:
\[
\rho(u_1) < \rho(u_2) < \cdots < \rho(u_k) < \rho(u_1),
\]
which simplifies to $\rho(u_1) < \rho(u_1)$, a contradiction. Therefore, $G$ is acyclic.
\end{myproof}

\subsection{Proof for Kirchhoff’s Current Law characteristic preservation theorem}
\label{subsec:appendix_kcl_loss}
\begin{mytheorem}
Let two sets of vectors be $\{\Vec{I}_{\Delta_1}, \Vec{I}_{\Delta_2}, \ldots, \Vec{I}_{\Delta_n}\}$ and $\{\Vec{I}_{\Delta_1'}, \Vec{I}_{\Delta_2'}, \ldots, \Vec{I}_{\Delta_n'}\}$ in $\mathbb{R}^d$, where $n < \|\mathcal{D}\|$ and $0 \leq \Delta_i, \Delta_i' \leq \|\mathcal{D}\|$ for all $i$ and $n < d$.
Then, there exists a non-trivial linear map $\phi: \mathbb{R}^d \to \mathbb{R}$ such that $\phi(\Vec{I}_{\Delta_i}) = \phi(\Vec{I}_{\Delta_i'})$ for all $i = 1, 2, \ldots, n$. 
\end{mytheorem}

\begin{myproof}

We define difference vectors as $\mathbf{v}_i = \Vec{I}_{\Delta_i} - \Vec{I}_{\Delta_i'}$. Then, we let $V = \text{span}(\{\mathbf{v}_1, \mathbf{v}_2, \ldots, \mathbf{v}_n\})$ be the subspace spanned by these difference vectors. We aim to construct a non-trivial linear map $\phi(\cdot)$ such that:

$$
\phi(\Vec{I}_{\Delta_i}) = \phi(\Vec{I}_{\Delta_i'}), \quad \forall i = 1, 2, \ldots, n.
$$

This is equivalent to finding a non-zero vector $\mathbf{w}$ in the orthogonal complement $V^\perp$ of $V$, since any $\mathbf{w} \in V^\perp$ satisfies:

$$
\mathbf{w} \cdot \mathbf{v}_i = 0, \quad \forall i,
$$
and we can define $\phi(\mathbf{x}) = \mathbf{w} \cdot \mathbf{x} +c$, where $c$ is any constant number.

The difference vectors $\{\mathbf{v}_1, \mathbf{v}_2, \ldots, \mathbf{v}_n\}$ lie in $\mathbb{R}^d$, and there are at most $n$ such vectors.
Since $n < d$, the subspace $V$ spanned by these vectors has dimension at most $n$, i.e., $\dim(V) \leq n$.
The orthogonal complement $V^\perp$ of $V$ in $\mathbb{R}^d$ satisfies:
$$
\dim(V^\perp) = d - \dim(V) \geq d - n > 0.
$$
Thus, $V^\perp$ is non-trivial, and there exist non-zero vectors $\mathbf{w} \in V^\perp$.

Since, any non-zero vector $\mathbf{w} \in V^\perp$ defines a linear map: $\phi(\mathbf{x}) = \mathbf{w} \cdot \mathbf{x} + \Vec{c}$, where $c$ is any constant number. For each $i$, the difference vectors $\mathbf{v}_i$ satisfy:
$$
\mathbf{w} \cdot \mathbf{v}_i = 0.
$$

Hence, we have:
$$
\phi(\Vec{I}_{\Delta_i}) - \phi(\Vec{I}_{\Delta_i'}) = \mathbf{w} \cdot (\Vec{I}_{\Delta_i} - \Vec{I}_{\Delta_i'}) = \mathbf{w} \cdot \mathbf{v}_i = 0.
$$

Therefore, $\phi(\Vec{I}_{\Delta_i}) = \phi(\Vec{I}_{\Delta_i'})$ for all $i$, and the map $\phi$ satisfies the conditions of the theorem.
\end{myproof}

\begin{mycorollary}
Suppose that for each $i$, the distance between the corresponding vectors with normalization is sufficiently small, i.e., 
$ 1 - \hat{\Vec{I}}_{\Delta_i}^{\top}\hat{\Vec{I}}_{\Delta_i^{\prime}} \leq \epsilon$.
A smaller $\epsilon$ makes constructing the desired map $\phi(\cdot)$ easier.
\end{mycorollary}

\begin{myproof}
The squared Euclidean distance between vectors satisfies:
$$
\begin{aligned}
\left\|\hat{\Vec{I}}_{\Delta_i} - \hat{\Vec{I}}_{\Delta_i'}\right\|_2^2 
&= \left\|\hat{\Vec{I}}_{\Delta_i}\right\|^2 + \left\|\hat{\Vec{I}}_{\Delta_i'}\right\|^2 - 2\hat{\Vec{I}}_{\Delta_i}^{\top}\hat{\Vec{I}}_{\Delta_i'}, \\
&= 2 - 2\hat{\Vec{I}}_{\Delta_i}^{\top}\hat{\Vec{I}}_{\Delta_i'},\\
&= 2(1 - \hat{\Vec{I}}_{\Delta_i}^{\top}\hat{\Vec{I}}_{\Delta_i^{\prime}}),\\
&\leq 2\epsilon.
\end{aligned}
$$
The first inequality can be easily achieve by normalization operation,  with $\left\|\hat{\Vec{I}}_{\Delta_i}\right\|, \left\|\hat{\Vec{I}}_{\Delta_i^{\prime}}\right\| \leq 1$.
Thus, each difference vector $\mathbf{v}_i = \hat{\Vec{I}}_{\Delta_i} - \hat{\Vec{I}}_{\Delta_i'}$ satisfies $\|\mathbf{v}_i\|_2 \leq \sqrt{2\epsilon}$.

Let $A \in \mathbb{R}^{d \times n}$ be the matrix with columns $\mathbf{v}_1, \ldots, \mathbf{v}_n$. Its Frobenius norm is bounded by:
$$
\|A\|_F = \sqrt{\sum_{i=1}^n \|\mathbf{v}_i\|_2^2} \leq \sqrt{2n\epsilon}.
$$

The singular value decomposition (SVD) of $A$ is:
$$
A = U \Sigma V^T,
$$
where $U \in \mathbb{R}^{d \times d}$ and $V \in \mathbb{R}^{n \times n}$ are orthogonal matrices,
$\Sigma \in \mathbb{R}^{d \times n}$ is a diagonal matrix containing the singular values $\sigma_1, \sigma_2, \ldots, \sigma_n$ of $A$.

The Frobenius norm of $A$ is related to its singular values:
$$
\|A\|_F = \sqrt{\sum_{i=1}^n \sigma_i^2}.
$$
Since $\|A\|_F \leq \sqrt{2n\epsilon}$, it follows that:
$$
\sum_{i=1}^n \sigma_i^2 \leq 2n \epsilon.
$$

The null space of $A^T$ corresponds to the orthogonal complement $V^\perp$. The dimension of the null space of $A^T$ is:
$$
\dim(\text{null}(A^T)) = d - \text{rank}(A).
$$
Since $\text{rank}(A) \leq n$, we have:
$$
\dim(\text{null}(A^T)) \geq d - n > 0.
$$
The size of the smallest singular value of $A$, denoted $\sigma_{\text{min}}$, determines the numerical stability of the null space.
If $\sigma_{\text{min}}$ is small, the null space of $A^T$ becomes larger.
From the bound on the singular values, we thus have the following:
$$
\sigma_{\text{min}} \leq \|A\|_F \leq \sqrt{2n\epsilon}.
$$

As $\epsilon$ decreases, the singular values of $A$ decrease, making the null space of $A^T$ more prominent and easier to find non-zero vectors $\mathbf{w} \in \text{null}(A^T) = V^\perp$.
This makes it easier to find a non-zero vector $\mathbf{w} \in V^\perp$ to define the Required non-trivial linear map $\phi(\cdot)$.

Thus, a smaller $\epsilon$ makes it easier to construct the desired map $\phi(\cdot)$.

\end{myproof}

\end{document}